\documentclass{scrartcl}
\usepackage[utf8]{inputenc}

\usepackage{graphicx}
\usepackage{upgreek}
\usepackage{algorithm2e}
\usepackage{listings}
\usepackage{amsmath}
\usepackage{amssymb}
\usepackage{gensymb}
\usepackage{textcomp} 
\usepackage{hyperref}

\begin{document}
\title{Feature engineering workflow for activity recognition from synchronized inertial measurement units}
\author{Andreas W. Kempa-Liehr\footnote{\url{a.kempa-liehr@auckland.ac.nz}} \and Jonty Oram 
\and Andrew Wong
\and Mark Finch 
\and Thor Besier
}

\date{Department of Engineering Science, The University of Auckland, New Zealand
\\
IMeasureU Ltd, Auckland, New Zealand\\
Vicon Motion Systems Ltd, Oxford, United Kingdom\\
Auckland Bioengineering Institute, The University of Auckland, New Zealand
}
\maketitle

\begin{abstract}
The ubiquitous availability of wearable sensors is responsible for driving the Internet-of-Things but is also making an impact on sport sciences and precision medicine.
While human activity recognition from smartphone data or other types of inertial measurement units (IMU) has evolved to one of the most prominent daily life examples of machine learning, the underlying process of time-series feature engineering still seems to be time-consuming. This lengthy process inhibits the development of IMU-based machine learning applications in sport science and precision medicine.
This contribution discusses a feature engineering workflow, which
automates the extraction of time-series feature
on based on the FRESH algorithm (FeatuRe Extraction
based on Scalable Hypothesis tests) to identify statistically
significant features from synchronized IMU sensors (IMeasureU Ltd,
NZ).
The feature engineering workflow has five main steps: time-series
engineering, automated time-series feature extraction, optimized
feature extraction, fitting of a specialized classifier, and deployment of
optimized machine learning pipeline.
The workflow is discussed for the case of a
user-specific running-walking classification, and the generalization to
a multi-user multi-activity classification is demonstrated.
\end{abstract}

\section{Introduction}
Human Activity Recognition (HAR)
is an active research area within the field of \emph{ubiquitous sensing}, which has applications in
medicine (monitoring exercise routines) and
sport (monitoring the potential for injuries and enhance athletes performance).
For a comprehensive overview on this topic refer to 
\cite{OscarLabrador2013_HAR}. 
Typically the design of HAR applications has to overcome the following challenges \cite{Moran2015_HAR}:
\begin{enumerate}
    \item Selection of the attributes to be measured.
\item Construction of a portable and unobtrusive data acquisition system.
\item Design of feature extraction and inference methods.
\item Automated adjustment to new users without the need for re-training the system.
\item Implementation in mobile devices meeting energy and processing requirements.
\item Collection of data under realistic conditions.
\end{enumerate}

In this contribution, we are discussing the automated engineering of time-series features (challenge 3) from two synchronized inertial measurement units as provided by IMeasureU's BlueThunder sensor \cite{WongVallabh2018_IMeasureU}. Each sensor records acceleration, angular velocity, and magnetic field in three spatial dimensions. 
Due to the availability of machine learning libraries like tsfresh
\cite{tsfresh} or hctsa \cite{FulcherJones2017_hctsa}, which automate
the extraction of time-series features for time-series classification
tasks \cite{Fulcher2018_TSA}, we are shifting our focus from the
engineering of time-series features to the engineering of time-series.
For this purpose, we are considering not only the 18 sensor
time-series from the two synchronized sensors but also 6 paired
time-series, which measure the differences between the axes of
different sensors.
A further focus of this contribution is the optimization of the feature extraction process for the
deployment of the machine learning pipeline (Sec.~\ref{sec:workflow}).
The workflow is discussed for the case of a
user-specific running-walking classification (Sec.~\ref{sec:Jonty}), and the
generalization to a multi-user multi-activity classification
(Sec.~\ref{sec:ClassificaThor}) is demonstrated. The paper closes with
a short discussion (Sec.~\ref{sec:discussion})

\section{Automated feature engineering workflow}
\label{sec:workflow}
The automated feature engineering workflow presented in this paper has
two foundations: The BlueThunder sensor from IMeasureU
Ltd. \cite{WongVallabh2018_IMeasureU} and the time-series feature
extraction library tsfresh \cite{tsfresh,ACML2016_WLBD}.

\subsection{Synchronized inertial measurement units}
\begin{figure}[t]
  \begin{center}
\begin{tabular}{ll}
\textbf{a}& \textbf{b}\\
\includegraphics[height=0.5\textwidth]{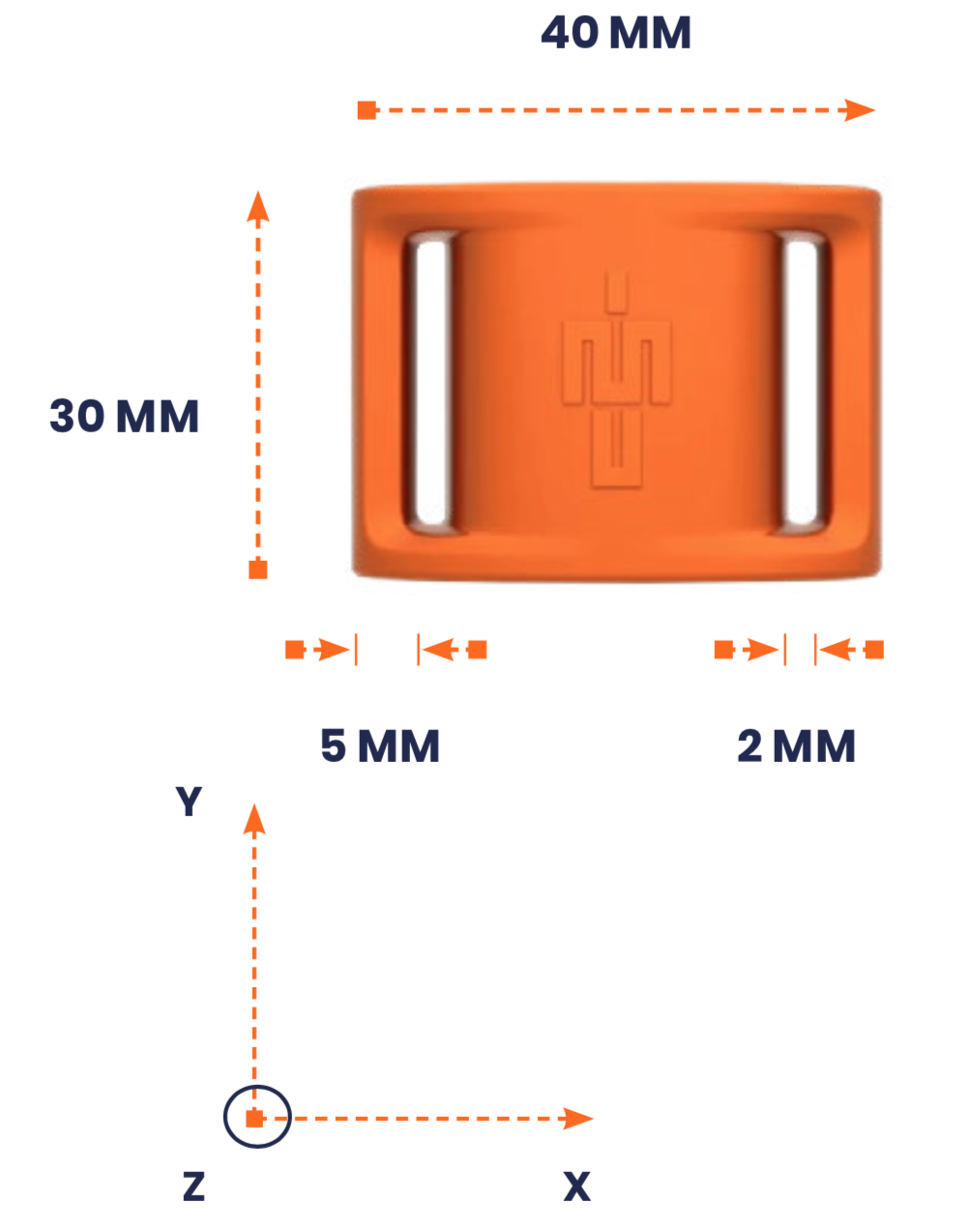} &
\includegraphics[height=0.5\textwidth]{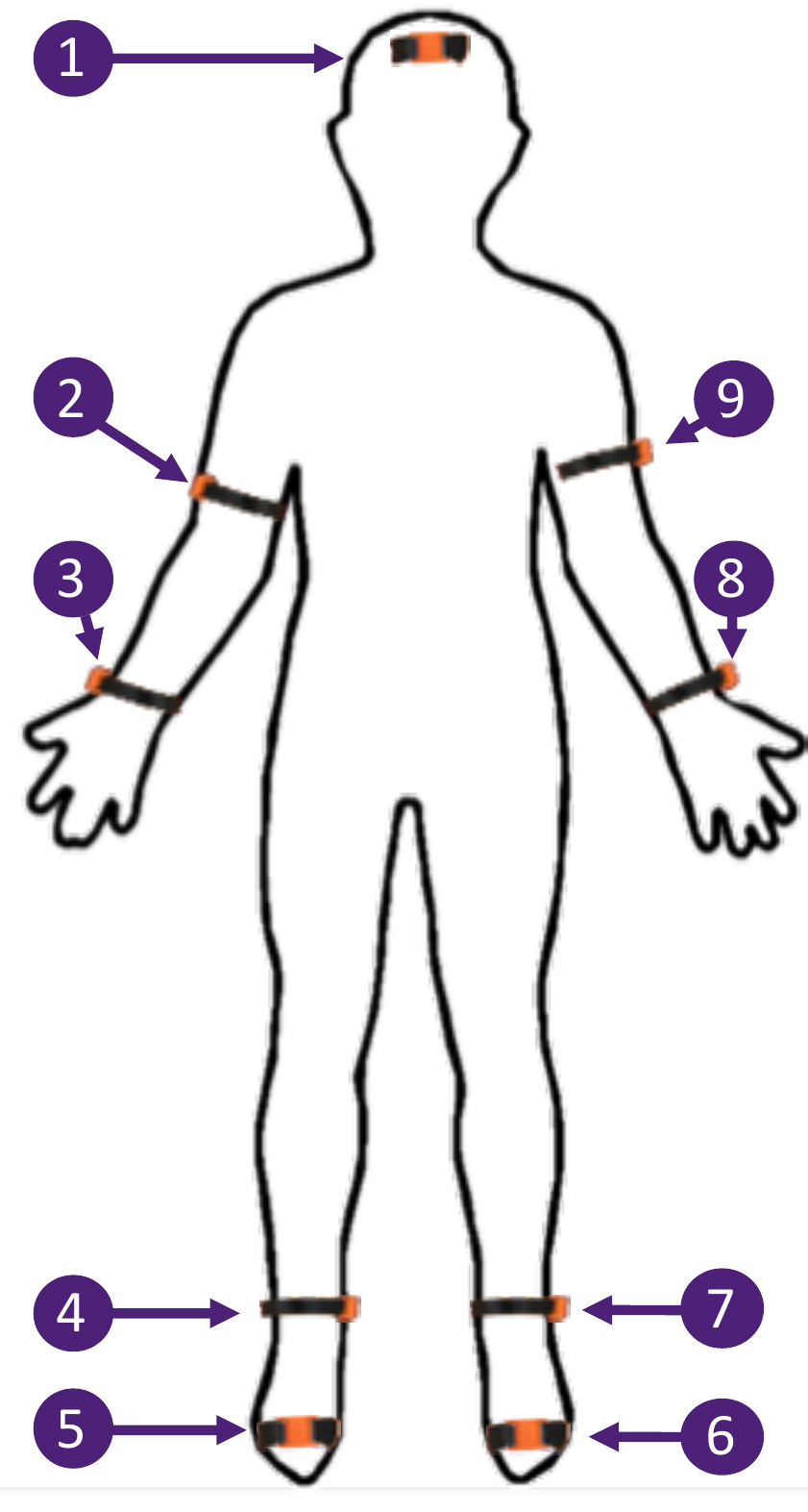}\\
\end{tabular}
\end{center}
\caption{IMeasureU's BlueThunder sensor. Panel \textbf{a} dimensions
  of sensor \cite[p.2]{WongVallabh2018_IMeasureU}, panel \textbf{b}
  Mounting points of sensors at the front of head (1), left and right upper arm
  (2, 9), left and right wrist (3, 8), left and right ankle (4, 7),
  and top of left and right foot (5, 6). For the running-walking
  classification, sensors were mounted at the left and right ankle
  (4, 7). For the multi-activity classification, the optimal sensor
  combination was tip of right foot (5) and right upper arm (2).}
\label{fig:sensor}
\end{figure}

The BlueThunder sensor is a wireless inertial measurement unit (IMU),
which combines a 3-axis accelerometer, a 3-axis gyroscope, and a
3-axis compass. Its specification is listed in Tab.~\ref{tab:IMU} and
its dimensions are shown in Fig.~\ref{fig:sensor}a.
One of the key features of this sensor is the fact that several units
can be synchronized. Therefore, not only the measured sensor signals itself,
but also paired signals, like, e.g. the difference between the
acceleration in the x-direction of two different sensors can be used
as an
additional signal.
One might interpret these computed signals as being recorded by
virtual sensors, which of course are basically signal
processing algorithms.

In order to demonstrate the applicability of the presented feature
engineering workflow, we are going to discuss two different activity
recognition experiments. The first experiment is concerned with the
discrimination of running vs walking for a specific person (Sec.~\ref{sec:Jonty}), the
second with generalizing the classification of 10 different activities
over different persons (Sec.~\ref{sec:ClassificaThor}).
The running vs walking classification experiment was designed
with a basic setup of two different IMUs being mounted at the left and
right ankle. The multi-activity classification task considered
9-different mounting points, which were mounted at the left and right
upper arm, the left and right wrist, the left and right ankle, as well
as the top of the left and right foot (Fig.~\ref{fig:sensor}b).

\begin{table}[t]
\caption{Specification of IMeasureU BlueThunder sensor
  \cite{WongVallabh2018_IMeasureU}.}
\label{tab:IMU}
\label{tab1e}
\begin{center}
\begin{tabular}{|l|l|}
\hline
\multicolumn{2}{|l|}{Features}\\
\hline
accelerometer range & $\pm16g$ \\
accelerometer resolution & 16 bit \\
gyroscope range & $\pm2000\degree{}/\text{s}$ \\
gyroscope resolution  & 16 bit \\
compass range & $\pm1200 \upmu\text{T}$ \\
compass resolution & 13 bit \\
data logging& 500Hz\\
weight & $12g$\\
\hline
\end{tabular}
\end{center}
\end{table}

\subsection{Feature extraction on the basis of scalable hypothesis testing}
At the core of the Python-based machine learning library
\texttt{tsfresh} \cite{tsfresh} is the FRESH algorithm. FRESH is the
abbreviation for \textit{FeatuRe Extraction on the basis of Scalable
  Hypothesis testing} \cite{ACML2016_WLBD}.
The general idea of this algorithm is to characterise each time-series
by applying a library of curated algorithms, which quantify each
time-series with respect to their distribution of values, correlation
properties, stationarity, entropy, and nonlinear time-series analysis.
Of course, this brute force feature extraction is computationally
expensive and has to be followed by a feature selection algorithm in
order to prevent overfitting. The feature selection is is done by
testing the statistical significance of each time-series feature for
predicting the target  and controlling the false discovery rate \cite{ACML2016_WLBD}.
Depending on the particular feature-target combination, the algorithm
chooses the type of hypothesis test to be performed and selects the
set of
statistically significant time-series features while preserving the
false discovery rate. The pseudocode of the FRESH algorithm is given
in Alg.~\ref{alg:FRESH} on p.~\pageref{alg:FRESH}.

\begin{algorithm}[b]
 \KwData{Labelled samples comprising different time-series}
 \KwResult{Relevant time-series features}
 \For{all predefined feature extraction algorithms}{
     \For{all time-series}{
       \For{all samples}{
              Apply feature extraction algorithm to time-series sample and compute time-series feature\;
      }
       Test statistical significance of feature for predicting the label\;
     }
     }
     Select significant features while preserving false discovery
     rate\;
     \caption{Pseudocode of Feature extRaction on the basis of Scalable
       Hypothesis testing (FRESH).}
     \label{alg:FRESH}
 \end{algorithm}

 \subsection{Feature engineering workflow for activity recognition}

 The general approach of the feature engineering workflow for activity
 recognition has five major steps:
  \begin{description}
\item[Time-series engineering] Increase the number of time-series by
  designing \textit{virtual sensors}, which combine the signals from
  different sensors, compute attributes like derivatives, or do both.
\item[Automated time-series feature extraction] Extract a huge variety of
  different time-series features, which are relevant for predicting the target.
\item[Optimized feature extraction] Identify a subset of features,
  which optimizes the performance of a cross-validated
  classifier. 
\item[Fitting of specialized classifier] Refit the classifier by using only the
  subset of features from the previous step.
\item[Deployment of optimized algorithm] Extract only those time
  series features,
  which are needed for the specialized classifier.
\end{description}

Note that the deployment step uses the fact that every feature can be
mapped to a combination of a specific time-series and a well-defined
algorithm.
Most likely, not all time-series are
relevant and depending on the classifier, only a small set of
time-series features is needed.
An example of this workflow is documented in the following case-study
for classifying running vs walking.

\section{Activity recognition case studies}
\label{sec:casestudies}
\subsection{Running vs walking}
\label{sec:Jonty}

The following case study trains an individualized activity recognition
algorithm for discriminating running vs walking on the basis of a
560 seconds long activity sequence, for which the corresponding
activities were logged manually:

\begin{itemize}
\item 2 synchronized IMUs mounted at left and right ankle (cf. Fig.~\ref{fig:sensor}b),
\item 560 seconds of mixed running and walking,
\item 280000 measurements for each of the 18 sensors (plus 6 paired measurements),
\item 140 sections of 4s length (82 walking-sections, 58 running-sections),
\item 15605 features in total,
\item 4850 statistically significant features (false discovery rate 5\%),
\end{itemize}

The virtual sensor was configured to compute the magnitude of
difference between corresponding directions of the acceleration and
gyroscope sensors. 
The time-series features were extracted with tsfresh
\cite{tsfresh}\footnote{\url{https://github.com/blue-yonder/tsfresh/tree/v0.10.1}},
which was available in version 0.10.1 at the time of this case study.
A random forest classifier as implemented in scikit-learn
\cite{pedregosa2011_scikitlearn} (version 0.19.0) was used for 
discriminating running vs walking.
The default configuration of the classifier already achieved 100\%
accuracy under 10-fold cross-validation, such that no hyperparameter
tuning was performed.
The following 20 time-series features were identified as optimized
time-series feature subset as features with the highest feature
importances from 100k fitted random forests.

\begin{lstlisting}[basicstyle=\scriptsize]
  ['accel_y_diff__agg_linear_trend__f_agg_"max"__chunk_len_5__attr_"stderr"',
  'accel_y_diff__change_quantiles__f_agg_"var"__isabs_True__qh_1.0__ql_0.0',
  'accel_y_r__agg_linear_trend__f_agg_"min"__chunk_len_10__attr_"stderr"',
  'accel_y_r__change_quantiles__f_agg_"mean"__isabs_True__qh_1.0__ql_0.0',
  'accel_y_r__change_quantiles__f_agg_"var"__isabs_False__qh_1.0__ql_0.2',
  'accel_y_r__change_quantiles__f_agg_"var"__isabs_False__qh_1.0__ql_0.4',
  'accel_z_diff__change_quantiles__f_agg_"var"__isabs_True__qh_1.0__ql_0.8',
  'accel_z_l__agg_linear_trend__f_agg_"min"__chunk_len_10__attr_"stderr"',
  'accel_z_l__change_quantiles__f_agg_"var"__isabs_False__qh_0.6__ql_0.0',
  'accel_z_r__minimum',
  'gyro_x_r__change_quantiles__f_agg_"var"__isabs_True__qh_0.4__ql_0.2',
  'gyro_y_diff__agg_linear_trend__f_agg_"max"__chunk_len_10__attr_"stderr"',
  'gyro_y_diff__agg_linear_trend__f_agg_"max"__chunk_len_50__attr_"stderr"',
  'gyro_y_diff__change_quantiles__f_agg_"var"__isabs_False__qh_1.0__ql_0.4',
  'gyro_y_diff__change_quantiles__f_agg_"var"__isabs_True__qh_1.0__ql_0.0',
  'gyro_y_l__change_quantiles__f_agg_"var"__isabs_True__qh_0.6__ql_0.4',
  'gyro_z_l__change_quantiles__f_agg_"var"__isabs_False__qh_0.6__ql_0.4',
  'gyro_z_r__change_quantiles__f_agg_"mean"__isabs_True__qh_0.6__ql_0.4',
  'gyro_z_r__change_quantiles__f_agg_"mean"__isabs_True__qh_0.8__ql_0.2',
  'gyro_z_r__change_quantiles__f_agg_"var"__isabs_False__qh_0.6__ql_0.0'] 
\end{lstlisting}

These 20 time-series features are computed from 10 different
time-series: four from the right ankle (\texttt{accel\_y\_r},
\texttt{accel\_z\_r}, \texttt{gyro\_x\_r}, \texttt{gyro\_z\_r}), three
from the left ankle (\texttt{accel\_z\_l}, \texttt{gyro\_y\_l},
\texttt{gyro\_z\_l}), and three magnitude of differences
(\texttt{accel\_y\_diff}, \texttt{accel\_z\_diff},
\texttt{giro\_y\_diff}).
Each feature references the generating algorithm using the following
scheme \cite{tsfresh}:
(1) the time-series \texttt{kind} the feature is based on,
(2) the name of the feature calculator, which has been used to extract the feature,
and
(3) key-value pairs of parameters configuring the respective feature calculator:
\begin{center}
  \lstinline{[kind]__[calculator]__[parameterA]_[valueA]__[parameterB]_[valueB]}
\end{center}
The features are dominated by two different methods, which quantify
the linear trend (\lstinline{agg_linear_trend}) and the expected
change of the signal (\lstinline{change_quantiles}). A detailed
description of the underlying algorithms can be found in the tsfresh
documentation\footnote{\url{https://tsfresh.readthedocs.io/en/v0.10.1/text/list_of_features.html}}. The
list of features can be converted into a dictionary using the
function
\begin{lstlisting}
tsfresh.feature_extraction.settings.from_columns
\end{lstlisting}
which can be
used for restricting the time-series feature extractor of tsfresh to
extract just this specific set of time-series features\footnote{\url{https://github.com/blue-yonder/tsfresh/blob/master/notebooks/the-fc_parameters-extraction-dictionary.ipynb}}.

\begin{figure}
  (a)\\
  \includegraphics[width=\textwidth]{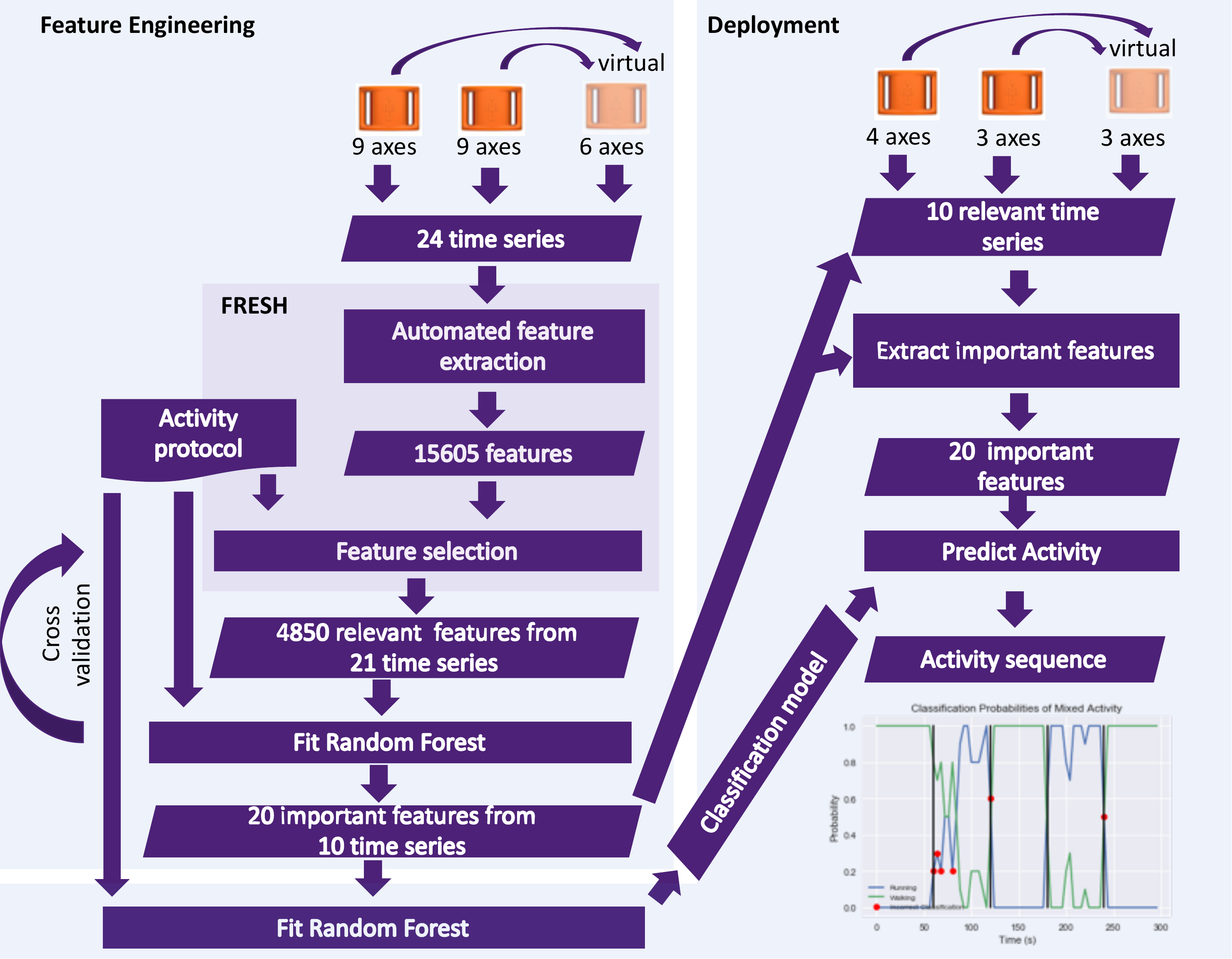}\\
  (b)\\
  \hspace*{0.1\linewidth}
    \includegraphics[width=.8\linewidth]{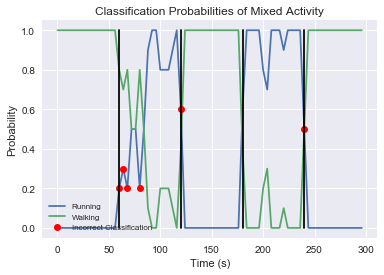}
  \caption{Feature engineering workflow for activity recognition
    tasks with details for the running vs walking case study. Classification of running vs walking for validation data set
operating on the 20 time-series features identified during the feature
engineering phase of the case study. Red dots indicate
misclassifications. The algorithm has an accuracy of 92\%.}
  \label{fig:workflow}
\end{figure}

Fig.~\ref{fig:workflow}a summarizes the feature engineering workflow
for the running vs walking case study. The inlay at the bottom right
of this figure is also depicted in Fig.~\ref{fig:workflow}b.
It shows the estimated activity sequence as time-series of
probabilities on a hold-out data set, which was recorded by the same
person as the training data set but on a different date.
For this activity classification, only the 20 time-series features
listed above were used. The algorithm's accuracy on the hold-out
dataset was 92\%.

\subsection{Multi-activity classification case study}
\label{sec:ClassificaThor}
The following case study involves a more complex feature engineering
setup because all nine sensor mounting points, as depicted in
Fig.~\ref{fig:sensor}, were considered for the feature engineering.
The task of this case study was to find a combination of sensors for
recognizing the activities
\begin{itemize}
\item laying down face down,
\item push-ups,
\item running,
\item sit-ups,
\item standing,
\item star jumps, and
\item walking,
\end{itemize}
while allowing for optimal generalization to other individuals.
Therefore, the feature engineering was optimized on the basis of a group
5-fold cross-validation of activities from five different persons (four
men, one woman). The mean accuracy for this proband-specific cross-validation
was 92.6\%.

\begin{figure}[t]
  \includegraphics[width=\textwidth]{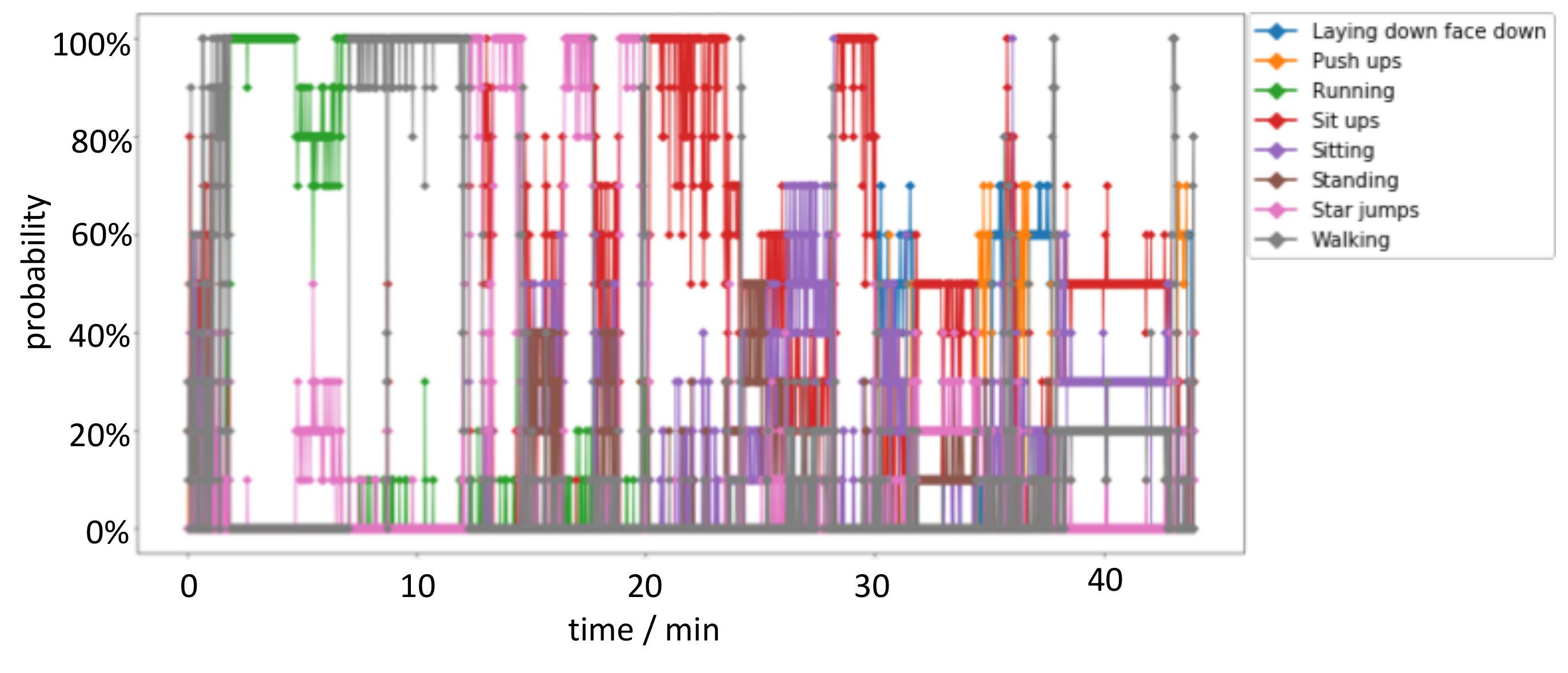}
  \caption{Evaluation of multi-activity recognition pipeline on the
    hold-out data set.}
  \label{fig:AndyEval}
\end{figure}

The optimal sensor mounting points for this task have been
identified as the tip of the right foot and the upper right arm
(Fig.~\ref{fig:sensor}).
The evaluation of the resulting activity recognition algorithm on a
sixth subject, who had recorded a 45min long evaluation data set,
retrieved a similar performance (Fig.~\ref{fig:AndyEval}) and was
computed in less than 20 seconds.

\section{Discussion}
\label{sec:discussion}
The presented workflow for feature engineering of activity recognition
task demonstrates a flexible and robust methodology, which is based
on the combination of signals from synchronized IMUs and automated
time-series feature extraction.
Due to the availability of machine learning libraries for automated time-series feature extraction, it can be expected that there will be a general shift of focus in research from the engineering of time-series features to the engineering of time-series. In this work, the engineering of time-series has been modelled as virtual sensors, but in many cases, this process will be similar to the design of signal operators.

\subsection*{Acknowledgement}
The authors like to thank Julie F\'{e}rard and the team at IMeasureU for
their support.


\end{document}